\documentclass[10pt,twocolumn,letterpaper]{article}

\usepackage[pagenumbers]{cvpr} % To force page numbers, e.g. for an arXiv version

\usepackage{graphicx}
\usepackage{amsmath}
\usepackage{amssymb}
\usepackage{booktabs}
\usepackage{multirow}
\usepackage{adjustbox}
\usepackage{sidecap}
\usepackage{float}

\usepackage{pifont}%
\usepackage[pagebackref,breaklinks,colorlinks]{hyperref}

\usepackage[dvipsnames]{xcolor}
\usepackage[capitalize]{cleveref}
\crefname{section}{Sec.}{Secs.}
\Crefname{section}{Section}{Sections}
\Crefname{table}{Table}{Tables}
\crefname{table}{Tab.}{Tabs.}

\def\cvprPaperID{694} % *** Enter the CVPR Paper ID here
\def\confName{CVPR}
\def\confYear{2023}

\newcommand{\cmark}{\ding{51}}%
\newcommand{\xmark}{\ding{55}}%

\newcommand{\Method}{CocoFormer}

\begin{document}

%%%%%%%%% TITLE - PLEASE UPDATE
%\title{Modeling Egocentric Object in the Wild for Visual Query Localization}
\title{Where is my Wallet? Modeling Object Proposal Sets\\ for Egocentric Visual Query Localization}

\author{Mengmeng Xu$^{1,2}\thanks{Work done during an internship at Meta AI.}$~~~
%{\tt\small mengmeng.xu@kaust.edu.sa}
\and
Yanghao Li$^{1}$~~~
\and
Cheng-Yang Fu$^{1}$~~~
\and
Bernard Ghanem$^{2}$
\and
Tao Xiang$^{1}$~~
\and
Juan-Manuel P\'erez-R\'ua$^{1}$
\and
%{\tt\small j.perez-rua@samsung.com}
{\small $^1$ Meta AI~~~~~~~~}
%{\small $^1$ SAIC Cambridge, UK}
% {\tt\small \{j.perez-rua,v.castillo,brais.a,xiatian.zhu\}@samsung.com}\\
{\small $^2$ KAUST, Saudi Arabia} \\
{\tt\small \{mengmeng.xu,bernard.ghanem\}@kaust.edu.sa,  } 
{\tt\small  \{lyttonhao,chengyangfu,txiang,jmpr\}@meta.com}\\
}

\maketitle

\begin{abstract}
%This paper deals with the problem of localizing objects in egocentric videos from provided visual queries. 
%In our setup, the query objects are defined by a tight visual crop that is ingested as input by our proposed system.
%Egocentric video recordings organically capture users' surroundings and their objects from non-conventional points of view. %This happens almost in an unintentional way,
%As daily-life objects are not always at the center of the user's attention they appear in more diverse locations, poses, and under varying levels of motion blur compared to third-view recordings. 
%The increased entropy of geometric transformations challenges the existing egocentric visual query systems, designed from common exocentric proposal-based object detectors, such as Faster-RCNN. 
%We found these unintentionally-recorded objects can be better modeled by a query-conditional contextual set transformer that is carefully trained with tailored data augmentations. 
This paper deals with the problem of  localizing objects in image and video datasets from visual exemplars. In particular, we focus on the challenging problem of egocentric visual query localization. We first identify grave implicit biases in current query-conditioned model design and visual query datasets. Then, we directly tackle such biases at both frame and object set levels. Concretely, our method solves these issues by expanding limited annotations and dynamically dropping object proposals during training. Additionally, we propose a novel transformer-based module that allows for object-proposal set context to be considered while incorporating query information. We name our module Conditioned Contextual Transformer or \Method.
%These simple but impactful techniques empower our transformer-based architecture design to exploit better global context from the available set of object proposals. We named this query-Conditioned Contextual Transformer \Method. 
Our experiments show the proposed adaptations improve egocentric query detection, leading to a better visual query localization system in both 2D and 3D configurations. Thus, we can improve frame-level detection performance from $26.28\%$ to $31.26\%$ in AP, which correspondingly improves the VQ2D and VQ3D localization scores by significant margins. %In particular, our model improves over the existing baseline solution for VQ2D by 5 points in stAP25. Furthermore, 
Our improved context-aware query object detector ranked first and second respectively in the VQ2D and VQ3D tasks in the 2nd Ego4D challenge. In addition to this, we showcase the relevance of our proposed model in the Few-Shot Detection (FSD) task, where we also achieve SOTA results. Our code is available at \url{https://github.com/facebookresearch/vq2d_cvpr}. 
% removing only for srt
\end{abstract}

%%%%%%%%% SECTIONS

\section{Introduction}
\begin{figure}[t]
\centering
\includegraphics[trim={9cm 0cm 10cm 0cm},width=8cm,clip]{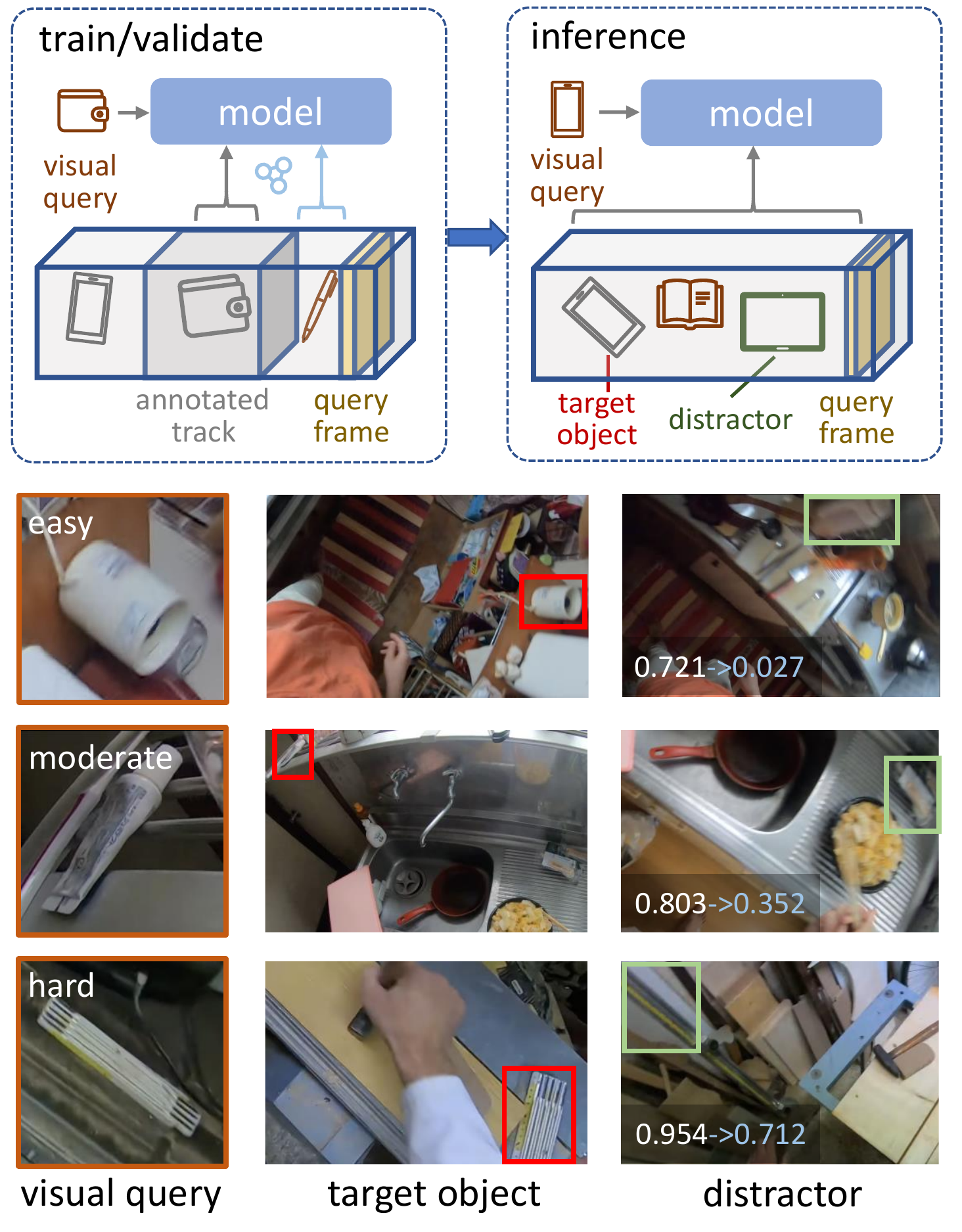}
\caption{
\textbf{Existing approaches to visual query localization are biased.} 
\textit{Top}: only positive frames containing query object are used to train and validate the query object detector, while the model is naturally tested on whole videos, where the query object is mostly absent. Training with background frames alleviates confusion caused by hard negatives during inference and helps predict fewer false positives in negative frames.
\textit{Bottom}: visual query, target object, and distractors of three query objects from easy to hard. The confidence scores in {\color{gray} white} from a biased model get suppressed in our model in {\color{cyan} cyan}. Our method improves the visual query localization system with fewer false positives, even for the very challenging scenario shown in the last row. 
}
\label{fig:intro}
\end{figure}

The task of Visual Queries Localization can be described as the question, `when was the last time that I saw X', where X is an object query represented by a visual crop. In the Ego4D~\cite{Ego4D2022CVPR} setting, this task aims to retrieve objects from an `episodic memory', supported by the recordings from an egocentric device, such as VR headsets or AR glasses. 
A real-world application of such functionality is localizing a user's items via a pre-registered object-centered image of them. 
A functional visual query localization system will allow users to find their belongings by a short re-play, or via a 3D arrow pointing to their real-world localization.

The current solutions to this problem~\cite{Ego4D2022CVPR,xu2022negative} rely on a so-called `Siam-detector' that is trained on annotated response tracks but tested on whole video sequences, as shown in Fig.~\ref{fig:intro} (top, {\textcolor{gray}{gray}} arrows). The Siam-detector model design allows the incorporation of query exemplars by 
independently comparing the query to all object proposals.
%implementing an \FX{independent} comparison function between queries and object proposals. 
During inference on a given video, the visual query is fixed, and the detector runs over all the frames in the egocentric video recording.

Although existing methods offer promising results in query object detection performance, it still suffers from \textit{domain} and \textit{task} biases. 
The \textit{domain bias} appears due to only training with frames with well-posed objects in clear view, while daily egocentric recordings are naturally out of focus, blurry, and undergoing uncommon view angles.
On the other hand, \textit{task bias} refers to the issue of the query object always being available during training, while in reality, it is absent during most of the test time. Therefore, baseline models predict false positives on distractors, especially when the query object is out of view. These issues are exacerbated by the current models' independent scoring of each object proposal, as the baseline models learn to give high scores to superficially similar objects while disregarding other proposals to reassess those scores. 

In Fig.~\ref{fig:intro} (bottom), we show the visual query, target object, and distractors for three data samples from easy to hard. The confidence scores in \textcolor{gray}{white} of the distracting objects are reported from a publicly-available `Siam-RCNN' model~\cite{Ego4D2022CVPR}. Due to random viewpoints and the large number of possible object classes that are exhibited in egocentric recordings, the target object is hard to discover and confused with high-confidence false positives. In this work, we show that these biases can be largely tackled by training a conditioned contextual detector with augmented object proposal sets. Our detector is based on a hypernetwork architecture that allows incorporating open-world visual queries, and a transformer-based design that permits context from other proposals to take part in the detection reasoning. We call this model Conditioned Contextual Transformer or~\Method. \Method~has a conditional projection layer that generates a transformation matrix from the query. This transformation is then applied to the proposal features to create query-conditioned proposal embeddings. These query-aware proposal embeddings are then fed to a set-transformer~\cite{lee2019set}, which effectively allows our model to utilize the global context of the corresponding frame. We show our proposed \Method~surpasses Siam-RCNN and is more flexible in different applications as a hyper-network, such as multimodal query object detection and few-shot object detection (FSD).

Our \Method~has the increased capability to model objects for visual query localization because it tackles the domain bias by incorporating conditional context, but it still suffers from task bias induced by the current training strategy. To alleviate this, we propose to sample proposal sets from both labeled and unlabeled video frames, which collects data closer to the real distribution. 
Concretely, we enlarge the positive query-frame training pairs by the natural view-point change of the same object in the view, while we create negative query-frame training pairs that incorporate objects from background frames. 
Essentially, we sample the proposal set from those frames in a balanced manner.
We collect all the objects in a frame as a set and decouple the task by two set-level questions: (1) does the query object exist? (2) what is the most similar object to the query? Since the training bias issue only exists in the first problem, we sample positive/negative \textit{proposal sets} to reduce it. 
Note that this is an independent process of frame sampling, as the domain bias impairs the understanding of objects in the egocentric view, while task bias implicitly hinders the precision of visual query detection.

Overall, our experiments show diversifying the object query and proposal sets, and leveraging global scene information can evidently improve query object detection. For example, training with unlabeled frames can improve the detection AP from $27.55\%$ to $28.74\%$, while optimizing the conditional detector's architecture can further push AP to $30.25\%$. 
Moreover, the visual query system can be evaluated in both 2D and 3D configurations. 
In VQ2D, we observe a significant improvement from baseline $0.13$ to $0.18$ on the leaderboard. In VQ3D, we show consistent improvement across all metrics. % is consistent among the metrics: Overall Success, Success*, L2, and Angle. 
In the end, our improved context-aware query object detector ranked first and second respectively in the VQ2D and VQ3D tasks in the 2nd Ego4D challenge, and achieve SOTA in Few-Shot Detection (FSD) on the COCO dataset.

\section{Related Work}
\noindent \textbf{Object detection}.
Deep learning models for object detection, either one-stage \cite{lin2017focal,centernet,zhou2019bottom,law2018cornernet,fcos,detr}, or two-stage models \cite{girshick2015fast,ren2015faster,he2015spatial,dai2016r} assume that a large amount of human-annotated data~\cite{coco,lvis,kuznetsova2020open} is available. Indeed, extending them to handle novel categories associated with little data is challenging. Effectively handling test-time exemplars on top of that is even  more so, and that is exactly what is required for visual query (VQ) tasks which are a form of the few-shot detection task.
% Can we give more connection to existing object detection methods?

\noindent \textbf{Few-shot detection and visual queries.}
In few-shot detection (FSD), we aim to produce an object detection model that can quickly adapt to novel object categories specified by a small dataset of per-class samples. 
One example is~\cite{kang2019few}, which encodes each novel class as a re-weighting vector that is applied to object proposal features. Similarly,~\cite{park2022hierarchical} proposes a query-support hierarchical comparison function in a contrasting learning regime which obtains outstanding results in the COCO FSD benchmark. A related task, focusing on incremental enrollment of novel categories while preserving the performance of old ones is known as incremental few-shot detection (iFSD)~\cite{perez2020incremental}. Both FSD and iFSD are similar to the visual query tasks in that detection is conditioned on a small exemplar set (the query in VQ, and the few shots in FSD). All these tasks suffer from an implicit bias in testing time that causes a large amount of false positives~\cite{Ego4D2022CVPR,xu2022negative,kang2019few,perez2020incremental,yin2022sylph, xu2019missing,Pardo_2021_CVPR}. This is, exemplar-based test-time annotations do not provide enough negative signals at both frame and proposal set levels. Additionally, it is naturally difficult for base class training in FSD and VQ to appropriately generalize to novel categories due to the limited data variability in both individual instances and the number of categories~\cite{yin2022sylph}. This results in low-accuracy predictions of the novel categories. 
In fact, these issues are exacerbated in egocentric visual query localization~\cite{Ego4D2022CVPR}. This is because egocentric video recordings organically capture users' surroundings and their objects from non-conventional points of view. This happens almost in an unintentional way, as daily-life objects are not always at the center of the user's attention they appear in more diverse locations, poses, and under varying levels of motion blur in comparison to third-view recordings. In our paper, we unify FSD and VQ tasks under a single framework, directly tackling these long-lived ailments.

\noindent \textbf{Detecting objects in hyper-networks}
Conditional neural processes~\cite{garnelo2018conditional} (CNP) are inspired by Gaussian processes' ability to quickly adapt a function to new data in test time. %In CNP, neural models replace human-defined priors as a source of knowledge to transfer to new tasks. 
In practice, CNP models learn to change the shape of a predictive function given input-annotation pairs for a target task. They are often leveraged for efficient (amortized training) few-shot learning for classification or regression tasks~\cite{garnelo2018conditional, ye2022contrastive}. On the other hand, hyper-networks~\cite{ha2016hypernetworks}, a closely-related framework, are often considered in adaptive forms of object detection like few-shot detection~\cite{wang2020frustratingly,han2021query,fan2020few,kang2019few,park2022hierarchical}, continual object detection~\cite{zhang2020class,joseph2021towards,doshi2020continual,perez2020incremental}, and single or multiple object tracking~\cite{zheng2021improving,xu2019joint,wang2019spm,zhu2018distractor,yu2020deformable,li2019gradnet,zhu2018online,bertinetto2016fully,hu2022siammask,zhang2022robust,yan2022towards,zhao2022tracking,ma2022unified,zhou2022global,tokmakov2021learning}. Hyper-networks predict model parameters of another model, providing an elegant mechanism for exemplar-conditioned model adaptation in~\cite{bertinetto2016fully,perez2020incremental,kang2019few,yin2022sylph,park2022hierarchical}. Concretely, their most instrumental contributions are the specific architectural designs that are tailored for the respective downstream tasks. These are, for example, the correlation filter in~\cite{bertinetto2016fully}, code generators in~\cite{perez2020incremental,yin2022sylph}, the re-weighting operator in~\cite{kang2019few}, and the hierarchical attention module in~\cite{park2022hierarchical}. In our work, we embrace the hypernetwork-based design and approach the task from an object proposal set perspective. Thus, we propose to use a task-tailored set model that naturally addresses some of the biases intrinsic to visual query and few-shot detection.

\noindent\textbf{Egocentric visual understanding and visual query.}
Although computer vision research has largely focused on third-person images and videos, and egocentric vision datasets are not as common across all major areas of research (image and video understanding, object detection, object tracking, \etc)~\cite{kinetics,ucf,caba2015activitynet,miech19howto100m,deng2009imagenet,lin2014microsoft,muller2018trackingnet,fan2019lasot,VOT_TPAMI,huang2019got,Soldan_2022_CVPR,wu2023NewsNet}, recent studies on egocentric vision showcase its own major challenges and often is linked to novel computer vision tasks~\cite{Damen2018EPICKITCHENS,Ego4D2022CVPR,ego-topo,sigurdsson2018charades,fathi2012social,edemekong2017activities,li2018eye,su2016detecting}.
%Egocentric vision offers special perspectives to the real world and gives more challenges. Unlike exocentric visual data~\cite{kinetics,gu2018ava,ucf,caba2015activitynet,miech19howto100m,deng2009imagenet,lin2014microsoft}, egocentric recordings~\cite{Damen2018EPICKITCHENS,Damen2020RESCALING,lee-cvpr2012,engagement,pirsiavash2012detecting,disney-social,pirsiavash2012detecting,lee-cvpr2012,disney-social} have less human bias, as they are passively captured from wearable cameras for a long, and often continuous period. The flexibility of the camera sometimes leads to unusual viewpoints, loss of focus, motion blur, and lacks temporal curation~\cite{Ego4D2022CVPR}. Therefore, there is a sizable domain gap between egocentric and exocentric vision.

One such task, visual query localization is proposed in Ego4D~\cite{Ego4D2022CVPR}. The idea is to have a super-human memory from an AI agent that can answer questions according to recorded visual experience. Specifically, the VQ task requires spatiotemporal localization of an object specified by a visual query (cropped image of the object). For this task, \cite{Ego4D2022CVPR} proposed to use a Siam-RCNN network, which extracts visual features~\cite{he2016deep} from the query crop and region proposals~\cite{ren2015faster} from the input frame. Subsequent classification based on the inner product of proposals and query features is used to discover the object that is most similar to the query. Our study shows the training protocol of Siam-RCNN is suboptimal due to multiple biases in data sampling, as well as the intrinsic lack of global context modeling that is shown by the aforementioned query-proposal classifier. In this paper, we specifically tackle these biases
% for several forms of the visual query task 
and demonstrate our effectiveness across several tasks.

% \subsection{Video localization}
% \FX{General context, then introduce remaining stages as the application of our detector, }

\section{Method}

\begin{figure*}[t]
\centering
\includegraphics[trim={1cm 3cm 5cm 6.5cm},width=0.95\linewidth,clip]{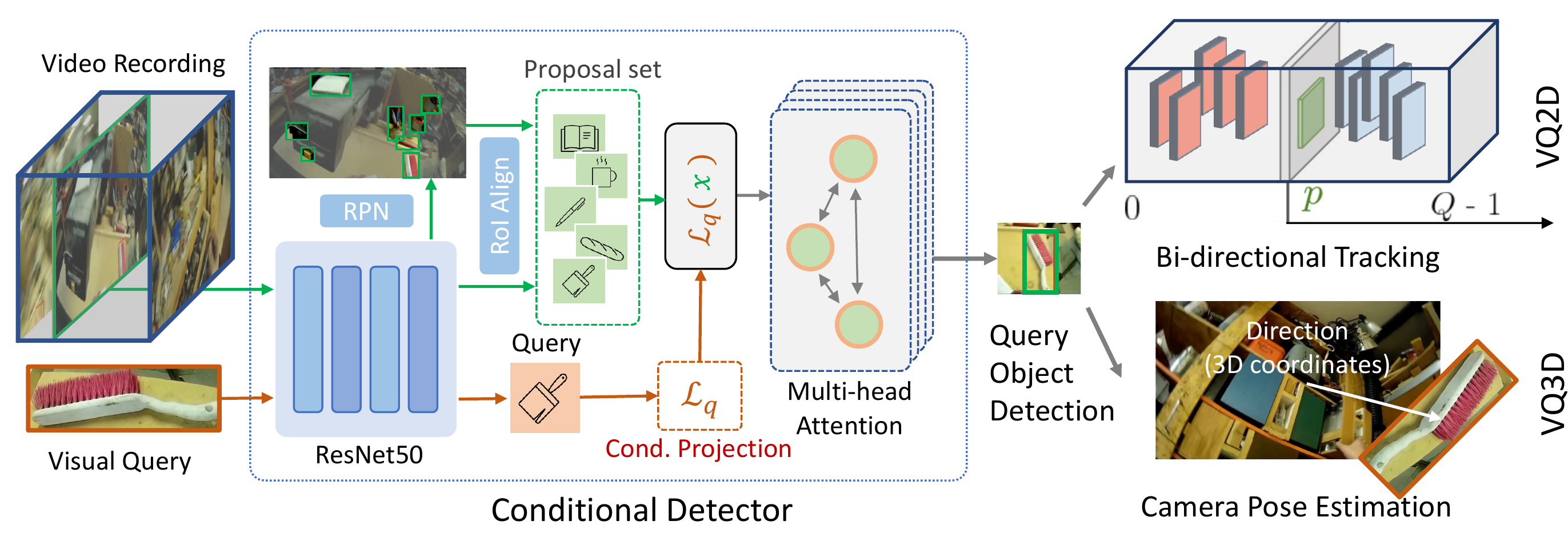}
\caption{
\textbf{Our detection + localization pipeline to solve
the VQ task.} 
We firstly \textbf{detect} the query object in the video recording conditioned on the visual query, then apply bi-directional tracking from the detection result to \textbf{localize} the object in the video (VQ2D), or use camera pose estimation to \textbf{localize} the predicted bounding box in real-world coordinates (VQ3D).
Specifically, our proposed \Method is trained with our augmented query-frame pairs, and tested on all the frames of video. 
It has a conditional projection layer that generates unique transformation from the query, and a multi-head attention block on the query-based proposal embedding to exploit global context. 
} 
\label{fig:pipeline}
\end{figure*}

The visual query (VQ) task takes as input a static image of the query object, and also a video recording from an egocentric view. The expected output is the localization of the object when it is last seen in the video. Specifically, the visual query 2D localization task (VQ2D) requires the output response to be a 2D bounding box for each frame of the temporal segment where the target object last appears. Additionally, the visual query 3D localization task (VQ3D) also requires a 3D displacement vector pointing to the 3D bounding box of the target object from the current camera position. VQ applies when a user asks for the location of an object by showing one image example, such as, `Where is this [the picture of a wallet]?'.

\subsection{Overall solution}
We developed a detection + localization pipeline to solve the VQ task.
The process is summarized in Fig.~\ref{fig:pipeline}. 
Given a visual crop $v$ of a query object $o$, we detect $o$ based on $v$ in all the frames of the video recording, denoted as $f$, by a conditional detector. Thus, the detector's inputs are the visual query crop $v$ and the video frame $f$, and the output is a bounding box with a confidence score $(x, y, w, h, c)$ in this frame.  
More specifically, our detector uses Region Proposal Network (RPN) with a ResNet-50 visual backbone to generate bounding box proposals $\{b_1,\cdots, b_N\}$ from the input frame, followed by an RoI-Align operation to extract bounding box features $\{\mathcal{F}(b_1),\cdots, \mathcal{F}(b_N)\}$. On the other side, the visual crop also passes through the same backbone, and generates a unique transformation from the conditional projection layer. The transformation is applied to proposal features, producing query-aware proposal embeddings for the multi-head attention layer, which predicts the target object from the proposal set.

We validate our query object detection result by VQ tasks defined in Ego4D. We strictly follow the episodic memory baseline of both tasks to localize the query object in the video and in the real-world coordinate. For VQ2D, we run a bi-directional tracker from the most recent detection peak, and predict the spatial-temporal bounding box track. For VQ3D, we apply camera pose estimation to the video frames, and output a displacement vector from the camera position to the predicted object~\cite{mai2022estimating}.

\subsection{Conditional Contextual Transformer}

Existing methods for VQ2D~\cite{Ego4D2022CVPR} follow a simple object-query pairwise comparison strategy.
A drawback of such formulation is that independent comparisons to the query limits model understanding by forcing a decision on the basis of individual similarity alone, disregarding the whole proposal set. We directly tackle such model bias by incorporating global set-level context in the form of an adapted set-transformer~\cite{lee2019set} architecture. Our transformer-based model takes the query-proposal feature pairs %for the top k=128 
for the top proposals as a set, and uses self-attention to learn the interaction between the query and all the selected proposals. 
% In our formulation, the training target is to predict whether the set of proposals actually include the query object. 
\begin{figure}[h]
\centering
\vspace{-0.2cm}
\includegraphics[trim={6cm 1.7cm 7.5cm 6cm},width=8cm,clip]{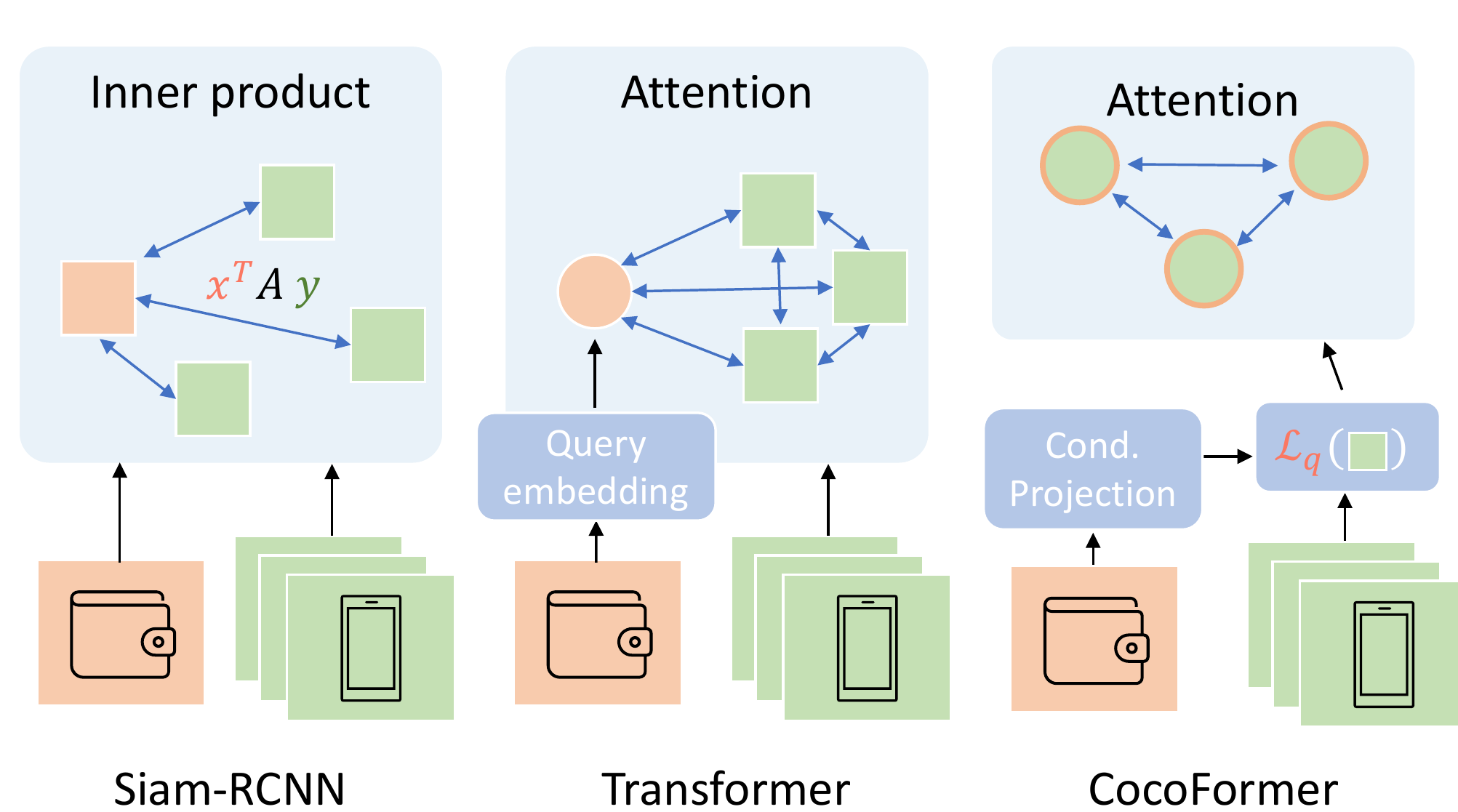}
\caption{
\textbf{We compare three model architectures for query object detection} 
\textit{Left}: The independent comparison in Siam-RCNN leads to the limited capability to model global context. 
\textit{Mid}: Directly using the attention model can hardly compare each proposal with the visual query. 
\textit{Right}: \Method~applies self-attention on the query-conditioned embeddings, thus the proposal selection is based on the query, and it exploits global context. 
}
\label{fig:method}
\end{figure}

We denominate our model Conditional Contextual Transformer or \Method~for short, since it provides set-level context and it is conditioned on the provided query. Concretely, our \Method~contains a conditional projection layer to generate a query-dependent embedding for each proposal candidate, and a self-attention block to exploit global context from the available proposal set in the camera view. 
The conditional projection layer $\mathcal{L}_q(x)$ generate a transformation (e.g. $\mathcal{L}_q$) based on the given condition (query) $q$, then the condition-dependent transformation is apply to the input $x$ in parallel. Our transformation generator is implemented as a 3D tensor with size $(C_{out}, C_{in}, C_{cond})$, which is the dimensions of output, input, and condition, respectively. This tensor firstly operates with the query $q$ on the last dimension, producing a $C_{out}\times C_{in}$ linear transformation. Then the transformation is applied to the input $x$ to get the new embedding.
In practise, we take the feature of visual crop as the condition, $q=\mathcal{F}(v)$, and the set of proposal feature as input $x=\{\mathcal{F}(b_1),\cdots,\mathcal{F}(b_N)\}$. Therefore, the output of this layer is the query-bias proposal embedding, i.e., $\{\mathcal{L}_{\mathcal{F}(v)}{(\mathcal{F}(b_1))},\cdots,\mathcal{L}_{\mathcal{F}(v)}{(\mathcal{F}(b_N))}\}$.  Similar to meta-train in few-shot learning, only the transformation generation process is optimized in training.
A multi-head attention block is used to improve the representation of each query-bias proposal embedding through the attention mechanism several times in parallel. More concretely, key, query, and value are all set to be $\mathcal{L}_{\mathcal{F}(v)}{(\mathcal{F}(b_N))}$, and the feed-forward network is an FC followed by binary classification.

Fig.~\ref{fig:method} compares our \Method~with baseline and transformer variants. Baseline Siam-RCNN has a pair-wise inter-production to decide if each proposal is the target object. The independent comparison of all the pairs leads to the limited capability to model global context. Alternatively, directly applying cross-attention or self-attention can enrich contextual awareness, but the model can hardly learn to compare each proposal with the visual query. \Method~applies self-attention on the query-conditioned embedding of each proposal candidate, thus it has a better ability to select elements based on the query and exploit global context. Please see Sect.~\ref{sect:advance} for ablation experiment.

% \subsection{Sampling from unlabeled frames}
\subsection{Augmented balanced proposal set}
As the sparsely annotated frames are only a small portion of the full Ego4d dataset, we propose to train our \Method~with Balanced Proposal Sets (BPS) from extended pseudo annotations on Positive Unlabeled Frames Sampling (P-UFS) and Negative Unlabeled Frames Sampling (N-UFS). We show these augmented proposal sets can minimize the domain and task bias in the training process.

\begin{SCfigure}[1][t]
\centering
\caption{\textbf{We augment positive pairs from positive unlabeled frames sampling (P-UFS).} The camera movement in the egocentric video creates different viewpoints, changing object positions, colors, brightness \etc.  
}
\vspace{0.3cm}
\includegraphics[trim={9.5cm 4.5cm 10.2cm 4cm},width=4.5cm,clip]{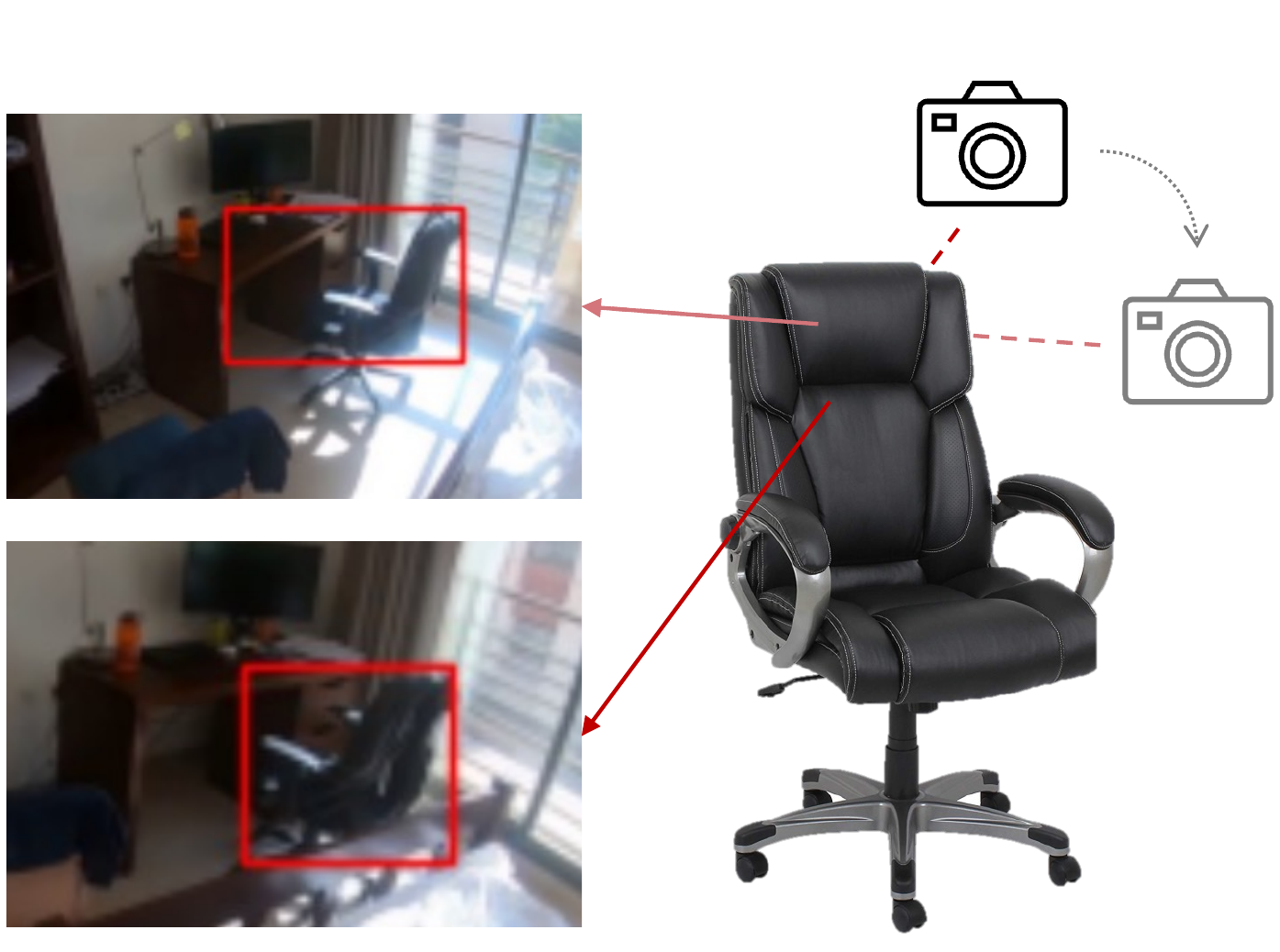}
\label{fig:vpj}
\end{SCfigure}

\noindent\textbf{Positive Unlabeled Frames Sampling}.
The space of all possible object instance appearances is a lot larger than existing annotations in real daily activities. To tackle this data deficiency, it becomes necessary to discover new objects in diverse viewpoints from P-UFS. 
Concretely, we run a vanilla pre-trained object detector on training videos, then take the resulting high-confidence detections and simulate query instances from them. Afterwards, we run a tracker in bi-direction to obtain different views of the same object, as shown in Fig.~\ref{fig:vpj}. This step can efficiently boost the object instances and effectively exploit the full visual information in training videos. 
% We design a heuristic algorithm to filter out unreliable samples, and finally include more positive samples and query objects during training. 
Please see Sect.~\ref{sect:sample} for the significance of these extra positive query-frame pairs.

\noindent\textbf{Negative Unlabeled Frames Sampling}.
Annotated positive frames often contain clear depictions of the object, while unlabeled objects with abnormal viewpoints, motion blur, and loss of focus are often ignored during annotation. In practice, however, these noisy samples greatly affect and confuse the detector. Thus, we also sample a number of new frames after the target object disappears from the camera view, and name the process Negative Unlabeled Frames Sampling (N-UFS). Those negatives are naturally closer to the data distribution in test time, and we can guarantee that the query object does not exist.
Note that N-UFS is applied along standard cross-batch negative sampling from positive frames. It adds objects from random frames that have no annotation. The advantage of training the detector with both positive and negative frames is to be more robust when applied to all the frames in the evaluation process.

\noindent\textbf{Balanced Proposal Set}.
A strong task bias in the VQ problem is that the training proposal set always contains the target object. This is indeed an issue since, in practice, a majority of frames do not contain the target object. This is the main reason why non-contextual models for VQ tend to predict false positives with large confidence. Clearly, without an appropriate proposal set sampling regime, baseline models will learn a shortcut to the VQ problem. This is, identifying the object that is most similar to the query irrespective of actually being the target.
%Therefore, even if we develop a VQ detector with increased capability to model egocentric objects with the context in the view, it leads to the risk of learning a task short-cut. For example, a trivial biased solution can be reporting the most similar object to the query whenever the frame contains the object. 
To alleviate this issue, we propose to pose the VQ task as two sequential questions: (1) \textit{does the query object exist in the proposal set} (2)\textit{ what is the most similar proposal to the query object}.
Since the task bias is mostly in the first question, we train the model by randomly sampling positive/negative proposal sets from a given frame. Concretely, we sample the RPN predictions to collect possible objects in the frame. The positive and negative proposal sets respectively include or exclude the object proposals that overlap with the ground-truth object. In this way, the existence of the query object is independent of the sampled frame, allowing us to effectively minimize the distribution gap between training and testing sets.

\section{Experiment}
\subsection{Dataset, tasks, and evaluation metrics}
\noindent \textbf{Dataset}. The VQ annotation in the Ego4D dataset includes 433 hours of videos in 54 scenarios, and 22k visual queries of $~3k$ object classes with $\sim$8 frames for each query. 
% Different from the egocentric object detection dataset, the bounding box annotation in Ego4D is near-uniformly distributed across the image. The videos are split into disjoint sets for train/val/test settings. 
Only train/val annotations are open to the public, and the test set performance can only be evaluated on a restricted server. 

\noindent \textbf{Tasks}. We evaluate our method on three tasks: query object detection, visual query 2D localization (VQ2D), and visual query 3D localization (VQ3D). Query object detection requires a detector predicting the bounding box of the matching object in the given image, directly evaluating our detector at a frame level.%, by being biased by the fact that the query object always exists in the given frame. 
VQ2D task returns a spatio-temporal location consisting of a set of bounding boxes in contingent frames. VQ3D task gives a 3D displacement vector in the real-world coordinates pointed to the physical location of the query object from the camera at the query time. 

\noindent \textbf{Metrics}. We followed the metrics in Ego4D for each of the three tasks.
We use \{$AP$, $AP_{50}$, $AP_{75}$, $AR@10$\} for query object detection, \{$tAP_{25}$, $stAP_{25}$, $rec\%$, $Succ$\} for VQ2D, and \{$L2$, $angle$, $O.Succ$, $Succ*$\} for VQ3D. 

Please see \textit{supplementary material} for the details of metrics, implementation, and the FSD experiment setup.
\subsection{Compare with SOTA}

\noindent
\textbf{Query detection}.
Tab.~\ref{tab:det} compares our method with baseline detectors. Compared to~\cite{Ego4D2022CVPR}, exploiting P-UFS, N-UFS, and BPS improves the detector by reducing the false positive rate, and better representing query objects. 
%The last two rows show that Set Transformer architecture is more beneficial to query object detection than independently comparing the query object to each proposal. The best performance is achieved when extra object tracks and the new architecture are both applied to the baseline detector. 

\begin{table}[t]
\small
    \centering
    \caption{
\textbf{Our method surpasses the ego4d baseline and CVPR winner's method 2022 in the query object detection task.} The best performance is achieved when augmented training pairs and the new architecture are both applied.
% \YH{We can rename "Xu. \textit{etal}~\cite{xu2022negative}" to "challenge winner in CVPR-2022".
% Is "Baseline~\cite{Ego4D2022CVPR}" the same as "Siam-RCNN" in ablations like Table 4? If so, it's better to unify their names to avoid confusion.}
}
\vspace{-2mm}
    \begin{tabular}{c|c|c|c|c}
    \toprule
     method & $AP$ & $AP_{50}$ & $AP_{75}$ & $AR@10$  \\
      \midrule 
Siam-RCNN~\cite{Ego4D2022CVPR} & 22.51 &	44.68	 & 18.24 & 42.3 \\
Xu.~\etal$^\dagger$~\cite{xu2022negative} & {26.28} &	49.63	 & 23.91 & 46.8 \\
% Xu. \textit{etal} (reprod.) & {27.55} &	50.43	 & 26.16 & 47.3 \\
\midrule
% Improved Baseline (ours) & Neg. + Pos. & 28.74 &	52.25	 & 27.35 & 50.1 \\
% Set Transformer (ours) & Neg. &	30.35 & 57.03 & 24.30 & 45.3 \\
Our Method &	\textbf{31.26} &	\textbf{57.96}	 & \textbf{28.88} & \textbf{47.1} \\
\textit{gain} & (+18\%) & (+17\%) & (+21\%) &  (+1\%)\\
\bottomrule
\end{tabular}
    \begin{flushleft}
    \vspace{-3mm}
    $^\dagger$Challenge winner in CVPR-2022
    \vspace{-3mm}
    \end{flushleft}
\label{tab:det}
\end{table}

\noindent
\textbf{VQ2D localization}
We run our query detector on the target videos at FPS=5 and use the \textit{Kys} tracker~\cite{bhat2020know}  to generate final response tracks. Tab.~\ref{tab:vq2d} compares our method with the baseline and previous winner's methods on the validation set. A similar comparison on the test set can be found on the public leaderboard. Promisingly, the success rate has been improved to 48.37, which means nearly half of the predicted response tracks overlap with the ground truth. 

\begin{table}[ht]
    \centering
    \small
    \caption{
    \textbf{In VQ2D, we outperform the baseline and CVPR 2022 winner's method on the validation set and got the first rank on the test set.}
    Notably, we improve $Succ$ to 48.37, meaning nearly half of the predicted response tracks overlap with the ground truth.}
    \vspace{-2mm}
    \begin{tabular}{c|c|c|c|c}
    \toprule
      method  & $tAP_{25}$ & $stAP_{25}$ & $rec\%$ & $Succ$ \\
      % method  & tAP25$\uparrow$ & stAP25$\uparrow$ & rec\%$\uparrow$ & Succ$\uparrow$  \\
      \midrule 
    Siam-RCNN~\cite{Ego4D2022CVPR}  & 0.20 &	0.12 & 32.2 & 39.8\\
    Xu \etal$^\dagger$~\cite{xu2022negative}  & 0.22 & 0.15  & 35.29 & 43.07 \\ \midrule
    Our Method  & \textbf{0.27 } &	\textbf{0.20  } & \textbf{42.34 } & \textbf{48.37} \\
    \textit{gain} & (+22\%) & (+33\%) & (+20\%) &  (+12\%) \\
    \bottomrule
    \end{tabular}
    \begin{flushleft}
    \vspace{-3mm}
    $^\dagger$Challenge winner in CVPR-2022
    \vspace{-3mm}
    \end{flushleft}
    \label{tab:vq2d}
\end{table}

\noindent
\textbf{VQ3D localization}
We also follow the VQ3D framework to project the matching object into 3D world coordinates. The test set predictions are summarized in Tab.~\ref{tab:vq3d}. We observed a consistent improvement over most of the metrics, meaning our improvement can eventually help the 3D prediction. Note that the limited improvement is due to the missing sufficient real-world projections to apply our response tracks. We believe this could be fixed by a more advanced algorithm~(\eg~\cite{mai2022localizing}) to estimate camera pose in 3D.

\begin{table}[ht]
\small
\setlength{\tabcolsep}{4pt}
    \centering
    \caption{\textbf{In the VQ3D challenge, we obtain significant improvement over the baseline and get the second rank on the public leaderboard.} Note that we only optimize the detector model, and leave the baseline camera estimation model fixed.}
    \vspace{-2mm}
    \begin{tabular}{c|c|c|c|c}
    \toprule
      method  & $L2$ $\downarrow$& $angle$ $\downarrow$ & $O. Succ$ $\uparrow$ & $Succ*$ $\uparrow$ \\ % & QwP $\uparrow$ \\
      \midrule 
    Siam-RCNN~\cite{Ego4D2022CVPR} & 4.64 & 1.31 & 0.08 & 0.49  \\ %  & 0.16\\
    % iBL~\cite{xu2022negative} &	Neg & 0.22 & 0.15  & 35.29 & 43.07 \\  
    \midrule
    Our Method & \textbf{4.46} &	\textbf{1.23} & \textbf{0.09} & \textbf{0.51}  \\ % &  0.16\\
    \textit{gain}& (+3.8\%) & (+6.5\%) & (+13\%) &  (+4.1\%) \\
    \bottomrule
    \end{tabular}

    \label{tab:vq3d}
\end{table}

% \begin{figure*}[h]
% \centering
% \includegraphics[trim={0cm 0cm 0cm 0cm},width=13cm,clip]{img/vis.png}
% \caption{\textbf{Visualization a true-positive prediction (a-b) and a false-positive prediction (c-d).}
% \FX{add 2 or 4 more pairs.}}
% \label{fig:vis}
% \end{figure*}

\subsection{Advances of \Method}\label{sect:advance}
\Method~is a query-conditional contextual transformer containing a conditional projection layer to transform the proposal candidates based on the query object and multi-head self-attentions to operate the transformations.
% to be able to answer two questions: (1) does the target exist in the set, and (2) which element is most likely to be the target in the set. 

\begin{table*}[ht]
\small
    \caption{\textbf{\Method~works better than other methods in visual query localization.} \textit{Top}: Siam-RCNN is a strong baseline with our improved training strategy. \textit{Mid}: Transformer with cross-attention works slightly better than the baseline in detection AP and AP50, but the inaccurate localization ability (e.g. AP75) leads the suboptimal performance in VQ2D. \textit{Bottom}: The overall performance in \Method~is generally better than other designs, and the conditional projection layer could localize the target object more tidily in the image, and results in better 2D localization prediction.}
    \vspace{-2mm}
    \label{tab:arch}
    \centering
    \begin{tabular}{c|c|cccc|cccc}
    \toprule    
    \multirow{2}*{model}&query& \multicolumn{4}{c|}{visual query detection} & \multicolumn{4}{c}{visual query 2D localisation} \\
      &  operation & $AP$ & $AP_{50}$ & $AP_{75}$ & $AR@10$ & $tAP_{25}$ & $stAP_{25}$ & $rec\%$ & $Succ$ \\
      \midrule 
% SiamRCNN & Correlation. & 28.74 &	52.25	 & 27.35 & 50.1 & 0.253 & 0.190 & 40.29 & 47.70\\
Siam-RCNN & Inner Product & 28.74 &	52.25	 & 27.35 & 50.1 & 0.253 & 0.190 & 40.29 & 47.70\\ \midrule
Transformer & Self-Attention & 17.46 &	37.61	 & 13.01 & 39.7 & 0.148 & 0.007 & 26.86 & 30.18 \\
Transformer & Cross-Attention & 29.17 & 54.87 & 26.59 & 46.1 & 0.212 & 0.140 & 37.93 & 41.51\\ \midrule
\Method & Concatenation & \textbf{31.96} & \textbf{60.07} & {28.77} & \textbf{48.0} & 0.257 &  0.191 & 42.15 & 47.81\\ 
\Method & Conditional Projection &  31.26 &	57.96	 & \textbf{28.88} & 47.1 & \textbf{0.267} &	\textbf{0.195} & \textbf{42.34} & \textbf{48.37}\\

    \bottomrule
    \end{tabular}
\end{table*}

\noindent \textbf{Effect of model design}. 
We compare several possible query detection modules in Tab.~\ref{tab:arch} by applying the exact same training/validation protocol (including our proposed P-UFS, N-UFS, and BPS).
The baseline model Siam-RCNN is shown in the first row of Tab.~\ref{tab:arch}. It independently computes the similarity between the query and the candidate objects by the inner product. Siam-RCNN greatly benefits from our improved training strategy as can be seen by comparing to the query detection results reported in Tab.~\ref{tab:det}. 
We also compare against vanilla Transformer-based architectures in the second block of Tab.~\ref{tab:arch}. Self-Att means we take the visual query with a learnable embedding as an extra token in a self-attention layer, while Cross-Att means we use each candidate proposal as query, and the visual query as key and value in multi-head attention. Transformer with cross-attention works slightly better than the baseline in detection $AP$ and $AP_{50}$, but it underperforms in full VQ2D due to inaccurate localization in the strict setting (\eg $AP_{75}$).% leads the suboptimal performance in VQ2D.
The overall performance of \Method~is superior to the other designs. %In particular, the conditional projection variation of our model provides better VQ2D. 
An interesting observation from Tab.~\ref{tab:arch} is that $AP_{75}$ in query detection better correlates to VQ2D scores than other detection metrics. Our full~\Method~provides clear gains in the strict image-level query detection setting and across the board in VQ2D.

\noindent \textbf{Extension to multimodal queries}.\label{sect:text}
We further show that our \Method~is flexible when more object information in the form of text labels is given. Specifically, we take the CLIP~\cite{radford2021learning} embedding of the text prompt as ``a photo of a [object]'', where the `object' is replaced by the query's title when it is available. In Tab.~\ref{tab:text}, we concatenate the text embedding to the query embedding, apply one FC layer for dimension reduction, and feed the FC output to the conditional project layers. Modeling queries with multimodal information result in consistent performance improvement.

\begin{table}[h!]
    \caption{\textbf{Incorporating text in our visual query detector.} Our~\Method~is flexible enough to readily accept data from extra modalities, and successfully exploit text labels. }
\small
    \vspace{-2mm}
    \centering
    % \begin{adjustbox}{width=\columnwidth,center}
    \begin{tabular}{c|cc|ccc}
    \toprule
    
    Use & \multicolumn{2}{c|}{Query detection} & \multicolumn{3}{c}{VQ2D localization} \\
    
       Text? & AP & AP50 & tAP25 & stAP25 & Succ.  \\\midrule
  \xmark & 31.26 &	57.96	& 0.257 & 0.195 & 48.37 \\  
% before & True & 32.03 &	60.16	 & 28.57 & 48.3 \\
 \cmark &  \textbf{32.65} &\textbf{62.64} & \textbf{0.269} & \textbf{0.201} & \textbf{49.03}\\

    \bottomrule
    \end{tabular}
    % \end{adjustbox}
    \label{tab:text}
\end{table}

\noindent \textbf{Extension to few-shot object detection}\label{sect:fsod}
We adapt our hyper-network to a few-shot object detection framework~\cite{park2022hierarchical}, which proposes hierarchical attention and meta-contrastive learning. Note that we still need their Hierarchical Attention Module to encode spatial information as our input, and apply our \Method~ for detection. We do base-train of 1-shot, 3-shot, and 5-shot on the MS COCO dataset~\cite{coco}, and report the novel categories average precision in
Tab.~\ref{tab:fsod}. With our conditional contextual transformer, we can consistently improve the state-of-the-art. Please see the complete table in
our \textit{supplementary material}.

\begin{table}[h!]
\small
    \caption{\textbf{Assessing model performance (nAP) in Few-Shot Detection.} We show 1-shot, 3-shot, and 5-shot settings on MS COCO. We don't finetune the model on the novel split (\textit{novel ft.}) after base-train. Please see the complete table in our \textit{supplementary material}. $\dagger$ means reproduced result from QA-FewDet~\cite{han2021query}.}
    \centering
    \vspace{-2mm}
    % \begin{adjustbox}{width=\columnwidth,center}
    \begin{tabular}{l|c|ccc}
    \toprule   
    % \multirow{2}{*}{Method} & \multirow{2}{*}{novel ft.} & \multicolumn{3}{c|}{1-shot}& \multicolumn{3}{c|}{3-shot} & \multicolumn{3}{c}{5-shot} \\   
    Method & novel ft. & 1-shot & 3-shot & 5-shot \\
        % & & nAP & AP50 & AP75 & nAP & AP50 & AP75 & nAP & AP50 & AP75\\
 
\midrule
 TFA ~\cite{wang2020few} & \cmark & 3.4& 6.6& 8.3 \\ 
 CoRPN ~\cite{zhang2020cooperating} & \cmark & 4.1 & - & - \\ 
 FADI~\cite{cao2021few} &\cmark& 5.7 & - & -\\   
 Xiao \etal~\cite{xiao2020few} &\cmark&3.2&6.7&8.1 \\
 MPSR~\cite{wu2020multi} $\dagger$ &\cmark&2.3&5.2&6.7 \\
 % Fan \etal~\cite{fan2020few}  $\dagger$ &True&4.2&6.6&8.0 \\ 
 Zhang \etal~\cite{zhang2021hallucination} & \cmark&4.4&7.2&- \\
 % QA-FewDet~\cite{han2021query}&True &4.9&8.4&9.7 \\
 Meta-DETR~\cite{zhang2021meta} &\cmark& 7.5 & 13.5 & 15.4 \\ 
 DeFRCN~\cite{qiao2021defrcn} &\cmark&9.3&14.8&16.1\\
 \midrule
 Fan \etal~\cite{fan2020few}  $\dagger$&\xmark&4.0&5.9&6.9\\
 Meta FRCNN~\cite{han2022meta} & \xmark & 5.0 & - & - \\
 QA-FewDet~\cite{han2021query}&\xmark &5.1&8.6&9.5 \\
 FS-DETR~\cite{bulat2022fs} & \xmark &7.0&9.8&10.7\\
 DAnA~\cite{chen2021dual} &\xmark & 11.9 & 14.0& 14.4 \\  
 % DAnA$^\dagger$  &False& 11.1 & 21.8	& 10.4 &  13.8 & 26.8 & 13.0 & 13.8 &26.8 &13.0 \\ \midrule
 hANMCL~\cite{park2022hierarchical} &\xmark& 12.9 & 14.4& 14.5  \\  
 \textit{ours} & \xmark & \textbf{13.3} &\textbf{14.7} & \textbf{14.8} \\
    \bottomrule
    \end{tabular}
    % \end{adjustbox}
    \label{tab:fsod}
\end{table}

\subsection{Proposal set sampling from video}\label{sect:sample}
% We compare the proposal set sampling source and strategy to study the bias impact on our \Method. 

\noindent
\textbf{Unlabeled Frame Sampling (UFS) reduces domain gap}. The model is learned from limited annotated foreground frames by default. Those frames usually have a more clear view, but this is not always true in all the frames.
Fig.~\ref{fig:vpj_ablation} studies the absolute improvement
achieved by P-UFS with different sampling ratios. The number of positive training pairs is extended from 1.2 to 1.4, 1.7, 2.1, 2.6, and 4.2 million, and we observe that 1.7 million (43\% extra) pairs give the most gain. 
The first block in Tab.~\ref{tab:bias} further studies the proposal sets from N-UFS on Siam-RCNN, where the query object is absent. Although proposal sets from background frames can slightly reduce the detection performance, they help localize the query object slightly better in the VQ2D localization task. Those experiments prove the unlabeled object helps to reduce the domain bias through wider data distribution. Please see \textit{supplementary material} for more comparisons on both Siam-RCNN and \Method.

\begin{figure*}
\centering
% \vspace{-0.7cm}
\includegraphics[trim={0cm 6cm 0cm 0.5cm},width=0.97\textwidth,clip]{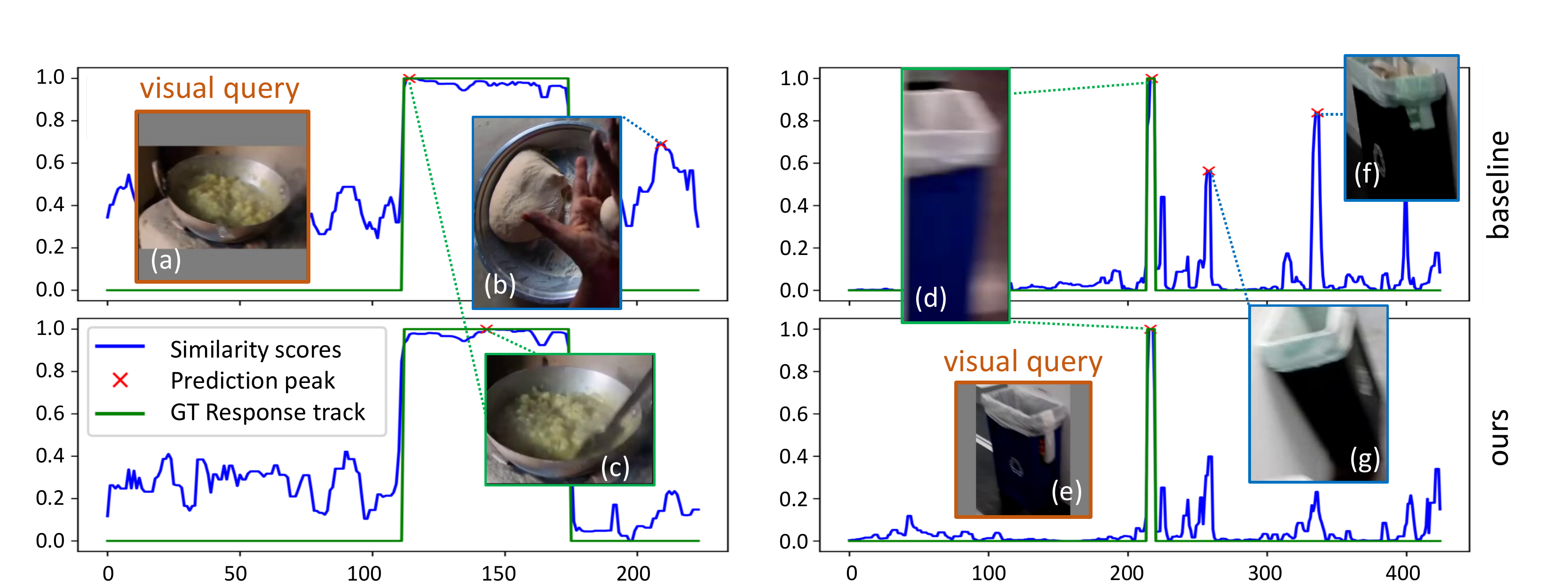}
\caption{\textbf{Visualization of our model \vs baseline.} The four plots show the predictions of two videos with baseline (top) and our method (bottom). Each data point on the similarity score curve indicates the confidence (y-axis) of the top-1 detected object at a video frame (x-axis). Although both methods discover the ground truths (c) and (d), the baseline method also reports false positives (b), (f), and (g). The videos can be viewed online at \href{https://visualize.ego4d-data.org/b2a1b8ca-99d6-4f26-953f-426e89649e90}{frying-pan video} from frame 120184, and \href{https://visualize.ego4d-data.org/b9b65293-fec8-494e-81e7-c4496aafba6f}{blue-bin video} from frame 1284.}
\label{fig:vis}
\end{figure*}
% \begin{table*}[ht]
% \small
%     \centering
%     \begin{tabular}{c|c|c| c c c c| c c c c}
%     \toprule
%     \multirow{2}*{model}& negative & negative & \multicolumn{4}{c|}{visual query detection} & \multicolumn{4}{c}{visual query 2D localisation} \\
%     & frame & object set & AP & AP50 & AP75 & AR@10 & tAP25 & stAP25 & rec\% & Succ \\
%       \midrule 
%     Siam-RCNN & False & False & {26.99} & 51.12 & 24.27 & 46.8  & 0.21 & 0.14 & 34.58 & 42.45\\
%     Siam-RCNN &  True & False & {26.28} & 49.63 & 23.91 & -  & 0.22 & 0.15 & 35.29 & 43.01 \\ \midrule
%     \Method & True & False & 33.28 &	62.51	 & 30.83 & 48.5 & 0.220 & 0.150 & 36.81 & 42.98 \\
%     \Method & True & True & 31.26 &	57.96	 & 28.88 & 47.1 & 0.267 & 0.195 & 42.34 & 48.37 \\
%     \bottomrule
%     \end{tabular}
%     \caption{\textbf{Dataset biases exist in both frame level and object-
% set level.} We apply negative pair synthesis with 50\% random ratio. \textit{Top block}: Training with background frames can help localise the query object slightly better in the video. \textit{Bottom block}: Training with a negative object set can evidentally reduce this training bias, and achieve a good VQ2D performance.}
%     \label{tab:bias}
% \end{table*}
% save space to one column
\begin{table}[ht]
\small
    \caption{\textbf{Dataset biases exist in both frame level and object-set level.} \textit{Top block}: Training Siam-RCNN baseline with N-UPS can help localize the query object slightly better in the video. \textit{Bottom block}: Training our CocoFormer with a BPS can reduce this training bias, and achieve a good VQ2D performance.}
    \vspace{-2mm}
    \centering
    \begin{tabular}{c|c| c c| c c c}
    \toprule
    N- & \multirow{2}{*}{BPS} & \multicolumn{2}{c|}{ VQ detection} & \multicolumn{3}{c}{VQ 2D localization} \\
    UFS &  & $AP$ & $AP_{50}$ & $tAP_{25}$ & $stAP_{25}$ & $Succ$ \\
      \midrule 
    \xmark & \xmark & {26.99} & 51.12 & 0.21 & 0.14 & 42.45\\
    \cmark & \xmark & {26.28} & 49.63 & 0.22 & 0.15 & 43.01 \\ \midrule
    \cmark & \xmark & 33.28 &	62.51 & 0.22 & 0.15 & 42.98 \\
    \cmark & \cmark & 31.26 &	57.96 & 0.27 & 0.20 & 48.37 \\
    % \cmark & \xmark & 33.28 &	62.51 & 0.220 & 0.150 & 42.98 \\
    % \cmark & \cmark & 31.26 &	57.96 & 0.267 & 0.195 & 48.37 \\
    \bottomrule
    \end{tabular}
    \label{tab:bias}
\end{table}

\begin{figure}[h]
\centering
\includegraphics[trim={0.0cm 0cm -0.5cm 0cm},width=8.2cm,clip]{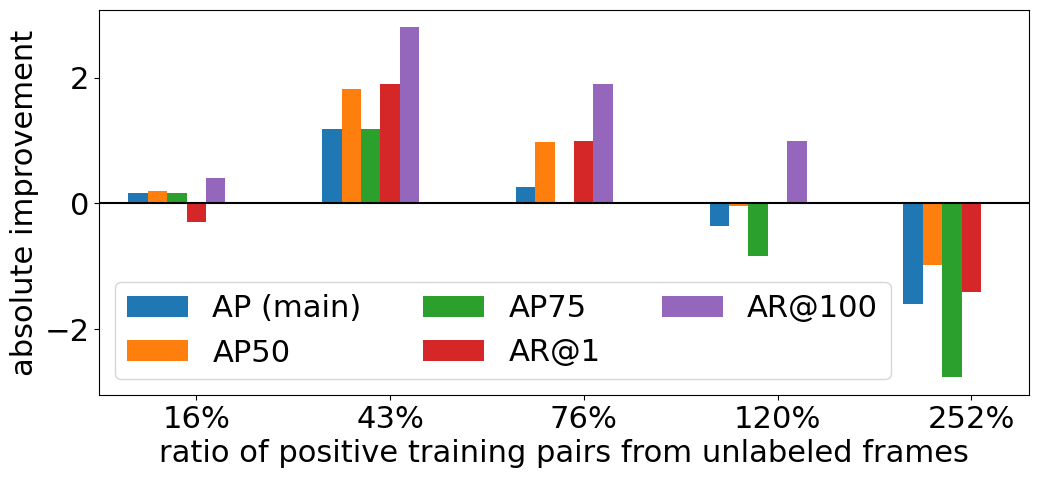}
\caption{
\textbf{We extend the positive training pairs and record the absolute improvement on different metrics.} The number of pairs gradually rises from 1.2 million up to 4.2 million. We observe that 1.7 million (43\% extra)  pairs give the most gains. It is also interesting to see AP75 drops faster when the pseudo pairs increase, as it requires more precise bounding-box predictions.
}
\label{fig:vpj_ablation}
\end{figure}

\noindent
\textbf{Balanced Proposal Set (BPS) reduces task bias}. 
% As a strong task bias in VQ, the query object always exists during training and validation, but it is absent in most of the video frames. 
% The existence bias will affect the model implicitly when the detector has individual comparison, and explicitly in \Method~because our prediction is from the proposal set. 
We study the efect of BPS and N-UFS on the VQ task bias in Tab.~\ref{tab:bias}.
The last two rows show the seriousness of the task bias issue.  BPS samples positive/negative proposal sets during training so that proposals overlapping with the ground truth are randomly removed. This simple strategy impairs the visual query detector performance (measured on positives frames only) as the proposal set does not always contain positives anymore, but it can evidently reduce VQ task bias, and achieve a much improved VQ2D performance.

\section{Qualitative Results}
% We visualize the predictions of two exemplar videos with baseline (top) and our method (bottom) in Fig.~\ref{fig:vis}. Each data point on the similarity score curve indicates the confidence (y-axis) of the top-1 detected object at a video frame (x-axis). The videos can be viewed online at \href{https://visualize.ego4d-data.org/b2a1b8ca-99d6-4f26-953f-426e89649e90}{frying-pan video} from frame 120184, and \href{https://visualize.ego4d-data.org/b9b65293-fec8-494e-81e7-c4496aafba6f}{blue-bin video} from frame 1284. In the frying-pan video, both methods find (c) is the same object as (a) in a different view. However, the baseline model also reports a high similarity score between query object (a) and distractor (b), as they are both metal containers with food material. 
We visualize the predictions of two exemplar videos in Fig.~\ref{fig:vis}.  In the frying-pan video, both methods find the query object in different views, but the baseline model also reports a high similarity score to the distractor, as it is also a metal container with food. 
The blue-bin video is more challenging because the query image crop was captured in a low-lighting condition. Therefore, the black bin detected by the baseline has a more similar visual appearance than the ground truth. Our method is able to model the global context and reports the black bin as negative.

\section{Conclusion}
% We study the challenging problem of egocentric visual query localization. 
We start by tackling Ego4D dataset and task biases of VQ2D, then propose to expand limited annotations and dynamically drop object proposals during training. Moreover, we proposed \Method, a novel transformer-based module that allows for object-proposal set context to be considered while incorporating query information.
Our experiments show the proposed adaptations improve egocentric query detection, leading to a better visual query localization system in both 2D and 3D configurations. 
\Method ranked first and second respectively in the VQ2D and VQ3D tasks in the 2nd Ego4D challenge, and achieve SOTA performance in the Few-Shot Detection task.

\noindent
\small
\textbf{Acknowledgments.} MX and BG are supported by the KAUST Office of Sponsored Research (OSR) through the VCC funding and SDAIA-KAUST Center of Excellence in Data Science and Artificial Intelligence (SDAIA-KAUST AI).

\setcounter{section}{0}
\renewcommand\thesection{\Alph{section}}
\section{Metrics and Implementation details}

\subsection{Metrics selection}

In each  task, we followed the metrics introduced in Ego4D~\cite{Ego4D2022CVPR}. 

\noindent\textbf{Query Object Detection}. We consider average precision (AP) as the main metric. It is the  precision averaged over different recalls of the multiple predictions on the image. 
We also compare $AP_{50}$/$AP_{75}$ to study the predicted bounding boxes on loose and tide criteria and the top-10 recall to study the missing detection problem. 

\noindent\textbf{VQ2D and VQ3D}.
Most of the metrics focus on the closeness of the prediction to the ground truth. $tAP_{25}$ and $stAP_{25}$ in VQ2D evaluate how closely in the temporal and spatio-temporal extent the predicted response track matches the ground truth, respectively, where the intersection over the union threshold is 0.25 by default. $L2$ and $angle$ in VQ3D measure the difference between the predicted and ground-truth displacement vectors in the real-world coordinates. For a fair reference, we also report success ($Succ$) and recovery percentage ($rec\%$) to study how many predictions overlap the ground-truth, and how many ground truths are discovered by predictions.

\subsection{Implementation details}

\noindent\textbf{Training details}.
Following the optimized VQ2D baseline~\cite{xu2022negative}, we implement our algorithms on Detectron2~\cite{wu2019detectron2}. The visual query detection is conducted on 4 8-V100 GPU nodes in a distributed machine learning cluster. Each experiment trains the detector for 125k iterations with an initial learning rate of 0.02, which decays at 50k and 100k iterations by 0.1. Our batch size is 64.

\noindent\textbf{Frame Sampling}.
The training frames are sampled from video when a response track annotation is available. 
Our negative unlabeled frame sampling (N-UFS) is based on a \textit{negative video} starting at the end of the response track until the query frame. We sample as many frames from this negative video as the number of positive frames.
When applying positive unlabeled frame sampling (P-UFS), we run a COCO-pretrained Faster-RCNN~\cite{ren2015faster} in on all training videos with FPS=1, and track~\cite{bhat2020know} the predicted object with a confidence threshold of $0.5$ on both forward and backward directions. We remove outliers of this object based on a pre-defined range of area and aspect ratio. In the optimal setting, we totally sample 1.7 million extra query-frame pairs to train the detector.

To achieve the visual query localization tasks, we apply our trained detector in the respective pipelines~\cite{Ego4D2022CVPR}. In VQ2D, we run a \textit{Kys}~\cite{bhat2020know} tracker from the detection peak to predict the response track. In VQ3D, we leverage our improved query detector for frames where camera pose information is available. Note that we do not further modify these stages to ensure a fair comparison.

\begin{table*}[t]
\small
    \centering
    
    % \begin{adjustbox}{width=\columnwidth,center}
    \begin{tabular}{l|c|ccc|ccc|ccc}
    \toprule   
    \multirow{2}{*}{Method} & \multirow{2}{*}{novel ft.} & \multicolumn{3}{c|}{1-shot}& \multicolumn{3}{c|}{3-shot} & \multicolumn{3}{c}{5-shot} \\    
        & & nAP & AP50 & AP75 & nAP & AP50 & AP75 & nAP & AP50 & AP75\\
 
\midrule
 TFA ~\cite{wang2020few} & True & 3.4 & 5.8 & 3.8 & 6.6&12.1&6.5& 8.3& 15.3 & 8.0 \\ 
 CoRPN ~\cite{zhang2020cooperating} & True & 4.1 & 7.2 & 4.4 & -&-&-& -& - & - \\ 
 Meta-DETR~\cite{zhang2021meta} &True& 7.5 &	12.5	& 7.7 & - & - & -& - & - & - \\ 
 FADI~\cite{cao2021few} &True& 5.7 &	10.4	& 6.0 & - & - & -& - & - & -\\   
 Xiao \etal~\cite{xiao2020few} &True&3.2&8.9&1.4&6.7&18.6&2.9&8.1&20.1&4.4 \\
 MPSR~\cite{wu2020multi} $\dagger$ &True&2.3&4.1&2.3&5.2&9.5&5.1&6.7&12.6&6.4 \\
 Fan \etal~\cite{fan2020few}  $\dagger$ &True&4.2&9.1&3.0&6.6&15.9&4.9&8.0&18.5&6.3 \\ 
 Zhang \etal~\cite{zhang2021hallucination} & True&4.4&7.5&4.9&7.2&13.3&7.4&-&-&- \\
 QA-FewDet~\cite{han2021query}&True &4.9&10.3&4.4&8.4&18.0&7.3&9.7&20.3&8.6 \\
 DeFRCN~\cite{qiao2021defrcn} &True&9.3&-&-&14.8&-&-&16.1&-&-\\
 \midrule
 Fan \etal~\cite{fan2020few}  $\dagger$&False&4.0&8.5&3.5&5.9&12.5&5.0&6.9&14.3&6.0 \\
 Meta Faster-RCNN~\cite{han2022meta} & False & 5.0 & 10.5 & 4.5 & -&-&-& -& - & - \\
 QA-FewDet~\cite{han2021query}&False &5.1&10.5&4.5&8.6&17.7&7.5&9.5&19.3&8.5 \\
 FS-DETR~\cite{bulat2022fs} & False &7.0&13.6&7.5&9.8&18.5&9.8&10.7&20.5&10.8\\
 DAnA~\cite{chen2021dual} &False & 11.9 & \textbf{25.6}	& 10.4 & 14.0 & \textbf{28.9} & 12.3 & 14.4 & \textbf{30.4} & 13.0 \\  
 % DAnA$^\dagger$  &False& 11.1 & 21.8	& 10.4 &  13.8 & 26.8 & 13.0 & 13.8 &26.8 &13.0 \\ \midrule
 hANMCL~\cite{park2022hierarchical} &False& 12.9 &	25.0	& 12.1 & 14.4& 28.0 & 13.3 & 14.5 &27.9 &13.3  \\  
 \textit{ours} & False & \textbf{13.3} &\textbf{25.6}&\textbf{12.6} &\textbf{14.7} & { \underline{28.8}} & \textbf{13.4} & \textbf{14.8} & \underline{28.9} & \textbf{13.6} \\
    \bottomrule
    \end{tabular}
    % \end{adjustbox}
    \caption{\textbf{Assessing model performance in Few-Shot Detection.} We show 1-shot, 3-shot, and 5-shot settings on the MS COCO dataset. nAP means the novel categories average precision. $^\dagger$ means reproduced result by QA-FewDet~\cite{han2021query}.}
    \label{tab:fsod}
\end{table*}

\section{Few-shot Object Detection}
\subsection{Experiment setup}
\noindent\textbf{Dataset}
Our few-shot object detection experiments are on the MS-COCO dataset~\cite{coco}. The novel/base splits follow the setting of Kang \etal~\cite{kang2019few}.
From the 80 object categories, we use the 20 classes that overlap with the PASCAL VOC~\cite{everingham2010pascal} dataset as novel classes and the remaining 60 as base classes. 
Similarly, 5000 images from the validation set are used for evaluation, while the rest images in training and validation sets are used for training.

\noindent\textbf{Training details}
Our few-shot object detection model follows the released Faster-RCNN design and training recipe in~\cite{park2022hierarchical}. Its Hierarchical Attention Module encodes spatial information in the object proposals, then we vectorize the enriched proposal representation and feed them to our CocoFormer. 
We do base-training for 1-shot, 3-shot, and 5-shot without fine-tuning. Each base training is independent and done on a single Tesla V100 machine for 12 epochs. 
The learning rate starts at 0.001 and increases by 0.1 times per 1000 steps. 
We used stochastic gradient descent to optimize the model with a momentum of 0.9 and a weight decay of 0.0001. 

\subsection{Full comparison with SOTA}
Tab.~\ref{tab:fsod} assesses model performance in Few-Shot Detection. 1-shot, 3-shot, and 5-shot settings are respectively applied on the MS COCO~\cite{caesar2018coco} dataset. 
We divide the methods into two groups. Methods in the first block require fine-tuning on the novel classes. Their models got further optimized on the support set, so the performance especially on higher shots is relatively higher. 
Our method belongs to the second group, where the model is directly evaluated after the base train. Comparing novel categories' average precision (nAP), our method can consistently improve the baseline~\cite{park2022hierarchical}, outperform state-of-the-art, and is competitive with the fine-tuning methods in the first block. Notably, our method achieves 13.3 nAP in 1-shot object detection, which shares a more similar problem setting as visual query object detection.

\subsection{Visual query \vs few-shot detection}
We would like to emphasize that although visual query  and few-shot detection share similar configurations, but they are identical to each other.

First, visual query detection is based on \textit{an instance-level dataset}, while few-shot detection is on the class level. This new task requires the system to localize exactly the same object registered by its visual crop. Therefore, more than one instance from the same classes can con-exist in the query video, but the metrics will penalize a wrong instance. For example, there are four bins in the blue bins video in the qualitative result, but we have to find the blue bin along the corner of the wall.  

Second, the \textit{episodic training strategy}, which is widely used in few-shot detection, is not the optimal solution in visual query detection. This is because we have only one visual crop of the query object and thousands of novel instances. Applying an episodic training strategy may slightly improve the model performance, but it will greatly increase the training time.
% \FX{Juan, is that what you mean in WhatsApp?}

\section{Supplementary experiment}

\noindent
\textbf{Siam-RCNN \vs CocoFormer}
Our CocoFormer and P-UFS improve the framework in different aspects. 
CocoFormer is a novel transformer-based module that allows for object-proposal set context to be considered while incorporating query information, while the main motivation of positive unlabeled frame sampling (P-UFS) is to reduce the training domain gap between the overall possible object instance and the existing annotations.

In Tab.~\ref{tab:vpj}, we further validate this simple augmentation method on the baseline detector and our proposed \Method. The comparisons in each block show our augmentation strategy P-UFS effectively extends the training set, bringing consistent performance gain in both settings. If we compare CocoFormer with Siam-RCNN with or without P-USF, we can find the AP score is improved, yet AR@10 becomes lower. This means CocoFormer is more strict about predicting positives, and the precision is greatly increased. 

\begin{table}[ht]
\small
    \centering
    \begin{tabular}{c|c|c|c|c|c}
    \toprule
     backbone & P-UFS & $AP$ & $AP_{50}$ & $AP_{75}$ & $AR@10$  \\
      \midrule 
Siam-RCNN & \xmark & 27.55 &	50.43	 & 26.16 & 47.3\\
Siam-RCNN & \cmark & \textbf{28.74} &	\textbf{52.25}	 & \textbf{27.35} & \textbf{50.1} \\ \midrule
\Method & \xmark & 30.35 &	57.87	 & 26.76 & 45.9 \\
\Method & \cmark & \textbf{31.26} &	\textbf{57.96}	 & \textbf{28.88} & \textbf{47.1} \\

    \bottomrule
    \end{tabular}
    \caption{\textbf{Our augmentation strategy effectively extends the training set.} We validate the augmentation on Siam-RCNN and \Method, and it shows consistent performance gain in both settings.}
    \label{tab:vpj}
\end{table}

\section{Further discussion}
Due to space limitations, we left some further discussion and insight in this section.

\noindent
\textbf{Performance mismatch between VQD and VQL}. Most of the experiment tables show the model performances are not consistent when evaluated on VQ detection and VQ localization, which means a top-performing detection model can be sub-optimal for temporal localization. This is mainly because VQD is only evaluated on \emph{individually annotated frames} of the dataset, while VQL is evaluated on the entire video. 
%Therefore, the performance mismatch is due to the different metrics over different data. 
% Our statistics show that 
Positive frames are on average only \emph{2\% of all the frames} in the video.
%, so the domain gap is not negligible. 
Also, VQD is heavily biased because annotated frames always contain the query object, while a randomly sampled video frame doesn't have this property. Thus, VQL is much more challenging than VQD.
% a biased model can easily achieve good performance in detection, but not in localization. 
In this paper, we presented both VQD and VQL metrics to \emph{prove} that a better detector doesn't always lead to a better localizer. This is precisely the main motivation for our work: to reduce training bias between VQD and VQL by introducing various sampling methods.

\noindent
\textbf{Concatenation and Conditional Projection} in our proposed CocoFormer are both \emph{possible settings}. Although Concatenation works better on VQD, Conditional Projection is generally better in VQL, showing that the tracking process in the localization model is more sensitive to AP75. It means %makes intuitive sense because the tracking process is initialized from the detected bounding box, while 
a precise bounding box is necessary to produce a correct response track.

\noindent
\textbf{N-UFS and BPS for VQL} follow our main idea to sample data close to the VQL \emph{real distribution}. From the detection perspective, these simple methods are nontrivial or even counterintuitive, as {clean images with the query object} are preferred.
% from the true distribution. The N-UFS frames can be blurry and out-of-focus, and half of the BPS samples do not have information about the target object, 
% so the detection performance decreases when they are applied. 
However, the real-world data in VQL is noisy and long-tailed, so we have to use % which makes the task different from VQD. 
N-UFS and BPS to create necessary samples in this domain, and we find they are quite effective.
% to reduce this domain gap. 
Both methods are harmful when evaluated on VQD but helpful and essential in VQL to suppress false positives, as shown by similarity scores on background frames in Fig. 5. % Note that our method gives little influence on the positive frames. 

%%%%%%%%% REFERENCES
{\small
\bibliographystyle{ieee_fullname}
\bibliography{egbib}
}

\end{document}